\documentclass[runningheads]{latex/llncs}
\makeatletter
\def\input@path{{latex/}}
\makeatother

\usepackage{eccv}

\usepackage{eccvabbrv}

\usepackage{graphicx}
\usepackage{booktabs}

\usepackage[accsupp]{axessibility}  

\usepackage{graphicx}
\usepackage{amsmath}
\usepackage{amssymb}
\usepackage{booktabs}
\usepackage{xcolor}
\definecolor{darkpink}{RGB}{230,50,150}   
\definecolor{darkgreen}{RGB}{0,150,0}   
\definecolor{darkyellow}{RGB}{204,153,0}
\usepackage{multirow}
\usepackage{float}
\usepackage{enumitem}
\usepackage[most]{tcolorbox}
\usepackage{xcolor}
\newtcolorbox{responsebox}{
    colback=orange!5!white,    
    colframe=orange!75!black,  
    fonttitle=\bfseries,
    arc=2pt,                   
    boxrule=0.8pt,             
    left=5pt, right=5pt, top=5pt, bottom=5pt
}

\usepackage{hyperref}

\usepackage{orcidlink}
\usepackage{comment}
\def\projname{EgoPHI\xspace}

\begin{document}

\title{EgoPHI: Estimating 3D Hand-Object Contact and Force from Egocentric Vision} 

\titlerunning{EgoPHI: Estimating Contact and Force from Egocentric Vision}

\author{Andela Ilic\orcidlink{0009-0009-4671-1186} \and
Rachel Schuchert\orcidlink{0009-0003-6005-9055} \and
Yijing Jiang\orcidlink{0000-0002-3680-9579} \and
Christian Holz\orcidlink{0000-0001-9655-9519}}

\authorrunning{A. Ilic et al.}

\institute{Department of Computer Science, ETH Zürich, Zürich, Switzerland\\[1em]
\small\href{https://siplab.org/projects/EgoPHI}{\color{magenta}{\texttt{https://siplab.org/projects/EgoPHI}}}}

\maketitle

\begin{figure*}[h]
\centering
\vspace{-2mm}
\includegraphics[width=\textwidth]{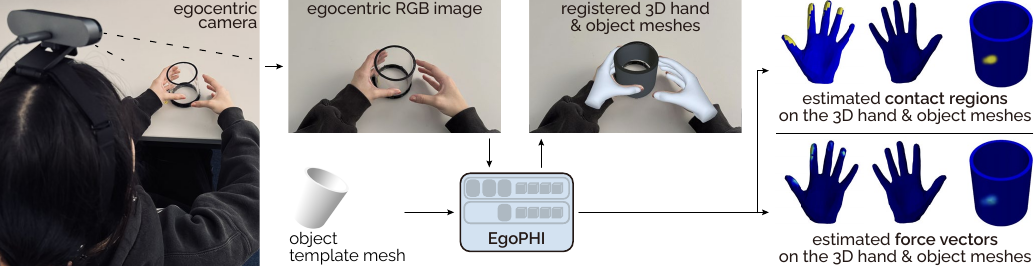}%
\vspace{-1mm}%
\caption{\projname is the first vision-based model that estimates 3D forces and contact during interaction with articulated rigid objects.
Given a single egocentric monocular RGB image and the object geometry, \projname registers hand meshes to refine the object pose in a shared camera coordinate frame, and predicts dense per-vertex contact and force distributions over all meshes for physically grounded interaction reasoning.}
\label{fig:teaser}
\vspace{-6mm}
\end{figure*}

\begin{abstract} 

Understanding hand-object interaction from egocentric vision is essential for modeling how people physically engage with the surrounding world.
Yet reasoning about physically grounded interaction requires estimating the forces acting on hands and objects, beyond localizing contact.
We present \emph{EgoPHI}, the first method that jointly estimates dense contact maps and 3D force distributions on hand and object meshes from a single monocular RGB image and object geometry.
To address the lack of scalable ground-truth force annotations, we introduce a physics-based simulation pipeline that augments existing hand-object datasets with dense per-vertex force supervision.
EgoPHI then learns dense 3D contact and force on interacting hand and articulated object meshes, extending vision-based force estimation beyond image-space or planar settings.
Our evaluation on in-distribution and out-of-distribution benchmarks shows that EgoPHI improves force estimation over existing approaches while generalizing to unseen datasets.
To evaluate sim-to-real transfer, we constructed two physical objects that capture dense object contact and force magnitude and used them to record a dataset of interactions from eight participants across diverse touch and grasp types.
Our results demonstrate that EgoPHI recovers meaningful 3D contact and force distributions in simulated, out-of-distribution, and real-world settings, advancing egocentric hand-object understanding from contact localization toward physically grounded interaction reasoning.
\\[.2em]
Code: \url{https://github.com/eth-siplab/EgoPHI}

\end{abstract}

\section{Introduction}
\label{sec:intro}

Understanding hand–object interaction is fundamental to modeling how humans engage with the physical world.
Detecting contact and estimating the magnitude and spatial distribution of forces during interaction reveal intent~\cite{8085141}, stability, and control~\cite{jiang2024contactimplicitmodelpredictivecontrol} in object manipulation.
For example, in robotics, recovering such information can enable robots to infer human manipulation strategies and learn stable grasping or object handling from observation.

Recent work has demonstrated predicting \emph{contact} from visual input, addressing both human–scene contact~\cite{huang2022capturinginferringdensefullbody,tripathi2023decodenseestimation3d,kim2024contactdiscoveringcomprehensiveaffordance,wang2025graceestimatinggeometrylevel3d,chen2023detectinghumanobjectcontactimages} and fine-grained hand–object contact~\cite{brahmbhatt2020contactposedatasetgraspsobject,hasson2019learningjointreconstructionhands,tse2023s2contactgraphbasednetwork3d,hu2024learningexplicitcontactimplicit,jung2025learningdensehandcontact,hampali2021ho3dv3improvingaccuracyhandobject}, where contact areas are small and vary during manipulation.
Aided by datasets~\cite{Taheri_2020,tripathi2023decodenseestimation3d} and benchmarks~\cite{hampali2021ho3dv3improvingaccuracyhandobject, chao2021dexycbbenchmarkcapturinghand}, methods now reliably detect contact points and regions even in challenging interactions~\cite{hasson2019learningjointreconstructionhands}.

However, contact alone is insufficient to fully capture the physical dynamics of hand–object interaction, which are ultimately governed by the \emph{forces} underlying these contacts.
Nevertheless, inferring forces from egocentric RGB is difficult due to occlusion, weak visual cues, and uncertain 3D geometry.
Subtle visual changes offer limited evidence of intensity, and single-view images provide little depth information to disambiguate 3D force direction.
Consequently, prior approaches have estimated forces only in the image space~\cite{grady2022pressurevisionestimatinghandpressure,grady2024pressurevisionestimatingfingertippressure} or flat surfaces~\cite{zhao2024egopressuredatasethandpressure}.

In this paper, we present \projname, a vision model that jointly estimates 3D forces and contact on both hands and the manipulated object from an egocentric view.
Taking a single egocentric image and the object geometry as input, our model leverages estimated hand poses to refine the object's 3D pose in a shared camera frame and infers dense contact and force distributions across all meshes.
For training \projname, we augment a large dataset of hand interactions with articulated rigid objects (ARCTIC~\cite{fan2023arcticdatasetdexterousbimanual}) with physically simulated contact forces during the recorded hand-object interactions.

Our evaluations on ARCTIC (in distribution) and H2O~\cite{kwon2021h2ohandsmanipulatingobjects} (out of distribution) show that \projname outperforms existing baselines in force accuracy and reduces the mean error by approximately 2\,N while generalizing to unseen objects.
To evaluate sim-to-real transfer, we captured a real-world hand–object interaction dataset with two fabricated objects (a cube and a cylinder) from eight participants.
We instrumented both objects to measure dense spatial contact and force distributions during touch and grasp.
Our evaluation provides the first experimental evidence that a vision-only system can estimate contact and 3D force distributions on non-planar real-world objects.

\vspace{.5em}
\noindent
\textbf{Contributions.}
We make the following contributions in this paper:
\begin{enumerate}[nosep]
\item \projname, the first vision-based method that estimates 3D force distributions and contact regions over both hand and object meshes during manipulation,
\item a physics-based force simulation module that generates realistic contact and force data from hand and object meshes, and
\item an instrumented real-world dataset of ground-truth hand-object contact and 3D forces, captured via two augmented objects from eight participants during touch and grasp interaction to validate \projname's sim-to-real transfer.
\end{enumerate}

\noindent
\projname establishes a basis for inferring physically grounded hand-object interaction dynamics from vision alone.
Our model performs robustly on unseen objects, diverse datasets, and real-world scenarios.
We believe this opens a new direction for studying hand-object manipulation in everyday settings.

\section{Related Work}
\label{sec:related_work}

\subsection{Vision-based contact area estimation}

Several projects have explored visual estimation of human–object and human–scene contact.
Previous work has targeted whole-body contact using parametric models (e.g., SMPL~\cite{SMPL:2015}, SMPL-X~\cite{SMPL-X:2019}) for dense 3D contact predictions across the full body, such as DECO~\cite{tripathi2023decodenseestimation3d}, PICO~\cite{cseke2025picoreconstructing3dpeople}, and GRACE~\cite{wang2025graceestimatinggeometrylevel3d}.
Others have aimed at fine-grained hand–object contact via MANO~\cite{MANO:SIGGRAPHASIA:2017}, such as S2Contact~\cite{tse2023s2contactgraphbasednetwork3d}, DyTact~\cite{cong2025dytactcapturingdynamiccontacts}, and Ti3D-contact~\cite{ti3dcontact2025}.
While single-image approaches can capture plausible contact for static frames, video-based methods such as VisTracker~\cite{xie2023visibilityawarehumanobjectinteraction} exploit motion cues to reason about contact dynamics.
HOT~\cite{chen2023detectinghumanobjectcontactimages} and PIHOT~\cite{wang2024precisionenhancedhumanobjectcontactdetection} predict 2D contact heatmaps in image space, which are useful for visual recognition but lack detailed geometric information.
Short of estimating contact masks, egocentric vision-based approaches have detected touch events on passive surfaces~\cite{streli2024touchinsightuncertaintyaware,streli2023structuredlightspeckle}.
Outside camera sensors, high-resolution contact masks have been estimated from low-resolution capacitive touch images using generative super-resolution approaches~\cite{streli2021capcontactsuperresolution} and foundation-model-based discriminative approaches~\cite{schuchert2026capfm}.
Conversely, methods that operate on meshes estimate dense 3D contact locations directly on the human or object surface (e.g., DECO~\cite{tripathi2023decodenseestimation3d}, PICO~\cite{cseke2025picoreconstructing3dpeople}, GRACE~\cite{wang2025graceestimatinggeometrylevel3d}).

In contrast to whole-body contact estimation, hand contact estimation is particularly challenging due to dual imbalance~\cite{jung2025learningdensehandcontact}.
Most hand vertices are not in contact during interaction, which creates a class imbalance.
In addition, spatial imbalance results from contact being predominantly concentrated on the fingertips, which limits accurate predictions in other hand regions.
The state-of-the-art (SOTA) model in this area, HACO~\cite{jung2025learningdensehandcontact}, addresses these challenges by introducing Balanced Contact Sampling (BCS) to mitigate class imbalance, and the Vertex-Level Class-Balanced (VCB) loss to alleviate spatial imbalance.
Existing datasets vary in the types of objects they contain.
They range from rigid objects~\cite{hasson2019learningjointreconstructionhands, chao2021dexycbbenchmarkcapturinghand, Cao2021, hampali2020honnotate, kwon2021h2ohandsmanipulatingobjects} to articulated objects~\cite{fan2023arcticdatasetdexterousbimanual, liu2024hoi4d4degocentricdataset} and deformable objects~\cite{xie2023hmdomarkerlessmultiviewhand}.

\subsection{Vision-based force estimation}

Early work by~\cite{ehsani2020useforcelukelearning}, \cite{Li_2019}, and \cite{7298898} demonstrated that forces and impulses can be inferred from object and human motion cues in RGB/RGB-D video.
These methods operate purely in image space without explicit 3D geometry~\cite{ehsani2020useforcelukelearning, Li_2019, 7298898}.
Similar ideas have been explored in robotics, where vision-based estimators inferred grasp or contact forces from RGB observations of robotic hands~\cite{Zhu2025ForcesFF}.
\cite{yang2024visionbasedforceestimationminimally} showed the feasibility of force estimation from visual deformation cues in minimally invasive surgery.
Recently, FORCE introduced a physics-guided dataset and model capturing human-applied forces and object resistance, which emphasized intuitive-physics priors for human–object interaction~\cite{zhang2024forcephysicsawarehumanobjectinteraction}.
ViTaM-D is a visual-tactile framework to reconstruct hand-object interaction via a force-aware contact representation~\cite{yu2025dynamicreconstructionhandobjectinteraction}.
Force information acts as a physical constraint to fine-tune the reconstruction derived from the visual input.

Closely related to our work are PressureVision~\cite{grady2022pressurevisionestimatinghandpressure}, PressureVision++~\cite{grady2024pressurevisionestimatingfingertippressure}, and EgoPressure~\cite{zhao2024egopressuredatasethandpressure}, which estimate pressure of hand contact with a flat surface from a single RGB image.
While PressureVision and PressureVision++ predict pressure as a 2D heatmap in the image space, EgoPressure shifts the problem to the egocentric perspective and focuses on estimating the pressure distribution directly on the 3D hand mesh.
Our work extends contact and force estimation from egocentric RGB images to both hands and articulated non-planar objects during interaction.
To our knowledge, we are the first to generalize vision-based force estimation beyond flat surfaces to include rigid and articulated objects.

\section{Method}
\label{sec:method}

\subsection{Problem statement}

Our objective is to estimate per-vertex contact points and interaction forces on the left and right hand meshes, denoted as $\mathcal{H}_L$ and $\mathcal{H}_R$, and on the object mesh $\mathcal{O}$ during hand–object interaction. 
To achieve this, \projname leverages both visual and geometric information: it takes as input a single egocentric RGB image $\mathcal{I}$, the 3D coordinates of the left and right hands $\mathcal{P}_L, \mathcal{P}_R \in \mathbb{R}^{V_H \times 3}$, and the 3D coordinates of the object $\mathcal{P}_O \in \mathbb{R}^{V_O \times 3}$, where $V_H$ and $V_O$ are the numbers of hand and object vertices, respectively.

\begin{figure*}[!t]
    \centering
    \includegraphics[width=\textwidth]{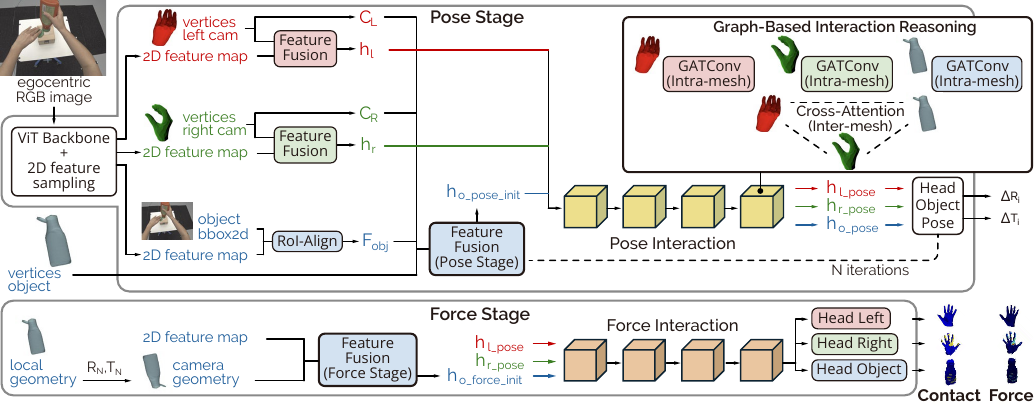}%
    \vspace{-1.25ex}%
    \caption{\projname's 3-stage pipeline: 
(1)~visual \& geometric feature extraction with cross-modal fusion, 
(2)~object pose estimation, and 
(3)~contact and force estimation. 
Our novel \emph{Graph-Based Interaction Blocks} perform the core 3D reasoning by encoding intra-mesh structure and inter-mesh relationships between the hands and object.
These blocks represent the 3D pose and spatial configuration of each mesh relative to the others.
We thus explicitly model proximity and geometric interaction, which is critical for inferring physically grounded forces.
}
    \label{fig:pipeline}
\end{figure*}

Hand pose estimation from a single image has become increasingly reliable in recent years~\cite{pavlakos2023reconstructinghands3dtransformers}.  
In contrast, object pose is often partially occluded, particularly in egocentric views where the hands frequently obstruct large portions of the object. 
To handle this, \projname estimates the object pose in the camera frame from a single egocentric RGB image $\mathcal{I}$, the object template mesh $\mathcal{O}_{\text{template}}$, and auxiliary cues derived from the image (object segmentation mask, hand centroids).
Given the inputs
\vspace{-1.0ex}
\[
\{ \mathcal{I}, \mathcal{O}_\text{template} \},
\vspace{-1.0ex}
\]
our goal is to predict per-vertex contacts
\vspace{-1.0ex}
\[
\mathcal{C}_L, \mathcal{C}_R, \mathcal{C}_O \in \mathbb{R}^{V_H \times 1}, \mathbb{R}^{V_H \times 1}, \mathbb{R}^{V_O \times 1}
\vspace{-1.0ex}
\]
and interaction forces
\vspace{-1.0ex}
\[
\mathcal{F}_L, \mathcal{F}_R, \mathcal{F}_O \in \mathbb{R}^{V_H \times 3}, \mathbb{R}^{V_H \times 3}, \mathbb{R}^{V_O \times 3}.
\]

\subsection{Method overview}

Figure~\ref{fig:pipeline} presents an overview of the \projname pipeline, which jointly estimates the object pose and per-vertex interaction forces and contacts. 
These force and contact values are obtained through physics-based simulation using the hand and object meshes available in the dataset.
Our pipeline consists of three main stages: \textbf{Visual and Geometric Feature Extraction}, \textbf{Iterative Pose Refinement}, and \textbf{Force Estimation}.
The first stage maps visual and geometric inputs into per-vertex features.
The second stage refines the object pose through successive updates.
The third stage then predicts per-vertex forces and contact areas on the final 3D object mesh.

\subsection{Physically Simulating Contact and Force}
\label{sec:simulation}

We supervise our model using physics-simulated, per-vertex contact and force distributions generated from existing hand–object datasets.

Given a dataset $\mathcal{D}$ comprising hand and object meshes and corresponding motion sequences, each frame includes the left and right hand meshes $\mathcal{H}_L$, $\mathcal{H}_R$, and the object mesh $\mathcal{O}$, all expressed in camera coordinates with vertex positions $\mathbf{v}_i \in \mathbb{R}^3$.
Our simulation then estimates a force vector $\mathbf{f}_i \in \mathbb{R}^3$ at every vertex $i$ on $\mathcal{H}_L$, $\mathcal{H}_R$, and $\mathcal{O}$.
This produces a dense per-vertex distribution of 3D forces
\vspace{-1.0ex}
\[
\mathcal{F} = \{\, \mathbf{f}_i \mid \mathbf{v}_i \in \mathcal{H}_L \cup \mathcal{H}_R \cup \mathcal{O} \,\}.
\vspace{-1.0ex}
\]

We provide $\mathcal{H}_L$, $\mathcal{H}_R$, and $\mathcal{O}$ aligned with their recorded poses to SOFA simulator~\cite{faure:hal-00681539}.
To estimate forces consistent with the observed hand-object configurations, we employ a penalty-based formulation that treats the meshes as quasi-rigid bodies.
We model interactions via soft virtual springs that activate when vertices enter a predefined contact proximity zone.
This formulation generates a restoring force $\mathbf{f}_i$ proportional to the collision displacement $d_i$,
\vspace{-2.0ex}
\[
\mathbf{f}_i = -k \, d_i \, \mathbf{n}_i
\vspace{-1.0ex}
\]
where $k$ represents the stiffness constant and $\mathbf{n}_i$ is the surface normal at vertex $i$.
Here, $d_i$ refers to the depth of the vertex within the contact threshold rather than a physical interpenetration of the underlying geometries.
Each mesh is imported as a triangle-based surface model with a uniform mass $m$.
We maintain a constant stiffness $k=10$ across all simulations, which provides stable and physically-consistent supervision for the rigid and articulated rigid objects.
Section A in the Appendix describes our implementation in more detail.

\begin{figure}[t]
    \centering
    \includegraphics[width=0.85\columnwidth]{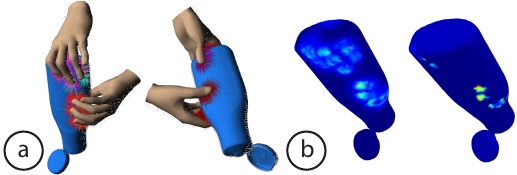}
    \caption{(a)~Activation of collision detection: \textcolor{darkpink}{pink} lines show line-line contact, \textcolor{blue}{blue} lines show point-line contact, \textcolor{darkyellow}{yellow} lines show point-point contact, and \textcolor{red}{red} lines represent constraints.
    (b)~Simulated forces before and after applying contact masks.\vspace{-.75ex}}
    \label{fig:force_correction}
\end{figure}

To derive contact areas on the meshes, we filter the raw forces $\mathcal{F}$ by a binary contact mask $\mathcal{M}$ that retains only physically valid contact regions.
The final ground-truth forces are therefore
\vspace{-.5ex}
\[
\tilde{\mathcal{F}} = \mathcal{M} \odot \mathcal{F},
\]
where $\odot$ denotes element-wise masking.
This step removes spurious non-contact forces and yields spatially consistent distributions over $\mathcal{H}_L$, $\mathcal{H}_R$, and $\mathcal{O}$ (Figure~\ref{fig:force_correction}).
The simulated forces and contact regions serve as dense, physically grounded supervision for \projname.
Combined with the ground-truth hand and object poses in the dataset, our augmentations provide realistic annotations to train our model's per-vertex contact and force prediction heads.

\subsection{Data preparation}
\label{sec:data_preparation}

Before training, we normalize all meshes in both local and camera coordinates by centering them at the origin and scaling them to unit magnitude.
We then map the articulated object mesh from its local coordinate system into the camera coordinate frame through a similarity transformation with rotation $R$ and translation $t$.
This aligns the template object to its ground-truth pose, and our pose-estimation model is later trained to predict these parameters $(R, t)$.

For the hand meshes $\mathcal{H}_L$ and $\mathcal{H}_R$ and the object mesh $\mathcal{O}$, vertex positions are expressed as $\mathbf{v}_i \in \mathbb{R}^3$ in camera space.
For force supervision, we normalize the simulated forces $\tilde{\mathcal{F}}$ by the maximum force magnitude observed across the dataset for the left hand, right hand, and object.
To improve numerical stability and training efficiency, we decompose each force vector $\mathbf{f}_i$ into its magnitude $|\mathbf{f}_i|$ and unit direction $\mathbf{u}_i = \mathbf{f}_i / |\mathbf{f}_i|$, rather than predicting the full 3D vector directly.
This representation enables cleaner loss formulations for force magnitude and direction and reduces sensitivity to extreme values in the raw force simulation.

\subsection{Model architecture}

\subsubsection{Visual and Geometric Feature Extraction}

Our model integrates a Vision Transformer (ViT) backbone to extract patch-level features from the input image.
The backbone returns a 2D feature map $F \in \mathbb{R}^{B \times C \times H_p \times W_p}$, where $B$ is the batch size, $C$ is the embedding dimension, and $H_p, W_p$ are the spatial dimensions of the patch grid.
We remove the classification head so that the feature map retains spatial detail needed for sampling features at projected mesh vertices.

To link visual cues with the 3D vertices of the hand meshes $\mathcal{H}_L$, $\mathcal{H}_R$ and the object mesh $\mathcal{O}$, we project each vertex $\mathbf{v}_i$ into the image using camera intrinsics.
We sample $F$ at the projected coordinates via bilinear interpolation and linearly embed the sampled vectors to the model's internal dimensionality~$D$.
Geometric and visual information jointly initialize the representation of each mesh vertex.

\noindent
\textbf{Hands:}
For each vertex $\mathbf{v}_i \in \mathcal{H}_L \cup \mathcal{H}_R$, the 3D camera-space coordinate is linearly projected into a $D$-dimensional space and concatenated with the corresponding sampled visual feature.
This concatenation is passed through a small multilayer perceptron (MLP) to produce the initial hand vertex embeddings.

\noindent
\textbf{Object~(Pose~Stage):} Our method dynamically computes the object's representation within an iterative pose refinement loop.
Each iteration integrates four distinct feature vectors:
(1)~\textbf{visual features} sampled from a RoI-aligned feature map $F_{\mathrm{obj}}$ using the current 2D projection;
(2)~\textbf{local geometry} from the template mesh coordinates $(x, y, z)$;
(3)~\textbf{relative geometry} to the left hand center;
(4)~\textbf{relative geometry} to the right hand center.
All four are embedded and combined to form the object vertex embeddings for the current iteration.

\subsubsection{Iterative Pose Refinement}

We model interactions between the left hand, right hand, and object using a stack of Graph-Based Interaction Blocks.
Each block contains three components:

\vspace{3pt}
\noindent
\textbf{1.~Intra-mesh Graph Attention Networks (GAT):}
These layers capture geometric relationships within each mesh ($\mathcal{H}_L$, $\mathcal{H}_R$, and $\mathcal{O}$).
Edges follow the mesh connectivity and the outputs of the GAT use residual connections and layer normalization.

\vspace{3pt}
\noindent
\textbf{2.~Inter-mesh Cross-Attention:}
Each mesh attends to the vertex features of the other two meshes (e.g., $\mathcal{H}_L$ attends to $\mathcal{H}_R$ and $\mathcal{O}$).
This way our model can capture inter-object interaction cues.

\vspace{3pt}
\noindent
\textbf{3.~Feed-forward Network (FFN):}
An MLP with a nonlinearity, residual connection, and layer normalization refines the vertex features.

\vspace{6pt}\noindent
We chain multiple Interaction Blocks to infer the object pose in two stages:

\vspace{3pt}\noindent
\textbf{Stage\,1.~Initialization:}
The object pose is initialized with rotation $R_0$ as the identity and translation $t_0$ as zero.

\vspace{3pt}\noindent
\textbf{Stage\,2.~Refinement Loop} (for $i=1$ to $N$):
\begin{itemize}[leftmargin=3.5mm,topsep=2pt]
\item \textbf{Feature Update:}
Given the current pose estimate $(R_{i-1}, t_{i-1})$, we update the object's vertex features while keeping the hand features fixed.

\item \textbf{Interaction:}
Static hand features and dynamic object features pass through the Interaction Blocks.

\item \textbf{Global Pooling:}
The resulting vertex features ($h_{l\_pose},\ h_{r\_pose},\ h_{o\_pose}$) are globally average-pooled and concatenated.

\item \textbf{Incremental Prediction:}
From the fused vector, a pose decoder predicts an incremental rotation $\Delta R_i \in \mathbb{R}^6$ (6D representation) and translation $\Delta t_i \in \mathbb{R}^3$.

\item \textbf{Pose Composition:}
The pose is updated by composing the increment: $R_i = \Delta R_i \cdot R_{i-1}$ and $t_i = \Delta t_i + t_{i-1}$.
\end{itemize}

\vspace{2pt}
\noindent
Through this iterative process, the model refines the object's pose by repeatedly re-evaluating the visual and geometric relationships across the three meshes.

\vspace{-8pt}
\subsubsection{Force Estimation}

We estimate per-vertex contact probability, normalized force magnitude, and direction in a second stage.
This stage is conditioned on the final predicted object pose $(R_N, t_N)$ from our refinement process.
We reuse the final hand features $(h_{l\_pose}, h_{r\_pose})$ from the last refinement iteration.
Object features are recomputed by transforming each object vertex $\mathbf{v}_i \in \mathcal{O}$ into camera space using $(R_N, t_N)$ and fusing two components:
(1)~visual features sampled from the full (non-RoI) backbone feature map, and
(2)~the final 3D camera-space coordinates of the object vertices.
These updated hand and object features are processed by a second and separate stack of Interaction Blocks dedicated to contact and force prediction.
The output heads for the left hand, right hand, and object predict per-vertex contact probabilities and force vectors.
Predicted forces are multiplied by the predicted contact probabilities to enforce physical consistency and to suppress non-contact forces.

\subsection{Loss Functions}
\label{sec:loss}
\vspace{-3pt}

We train our model with multiple objectives that jointly supervise contact prediction, force estimation, and object pose refinement.
Contact maps for $\mathcal{H}_L$, $\mathcal{H}_R$, and $\mathcal{O}$ are supervised with a weighted binary cross-entropy (BCE) loss.
We add three regularization terms to encourage spatial and semantic consistency~\cite{jung2025learningdensehandcontact}:
(i)~a vertex-level class-balanced (VCB) loss, 
(ii)~a dataset-level contact regularization term, and 
(iii)~a smoothness loss for locally coherent contact regions.

For every mesh vertex $\mathbf{v}_i$, we supervise both force magnitude and force direction.
The magnitude is trained with a contact-weighted L2 loss together with a regularization term that encourages physically reasonable mean magnitudes per mesh.
The direction is trained with a normalized L2 loss between predicted and ground-truth force vectors, weighted by the corresponding contact probabilities.

For pose estimation, we supervise the predicted pose using three terms:
(i)~a vertex reconstruction loss $\mathcal{L}_{\text{verts}}$, 
(ii)~a geodesic rotation loss $\mathcal{L}_{\text{rot}}$, and 
(iii)~a translation loss $\mathcal{L}_{\text{trans}}$.  
The full training objective combines all components:
\vspace{-1.0ex}
\begin{align*}
\mathcal{L}_{\text{total}} &= 
\lambda_c \mathcal{L}_{\text{contact}} + 
\lambda_m \mathcal{L}_{\text{force-mag}} + 
\lambda_v \mathcal{L}_{\text{force-vec}} \notag\\
&\quad + \lambda_t \mathcal{L}_{\text{trans}} + 
\lambda_r \mathcal{L}_{\text{rot}} + 
\lambda_p \mathcal{L}_{\text{verts}}
\vspace{-1.25ex}
\end{align*}
$\lambda_c, \lambda_m, \lambda_v, \lambda_r, \lambda_t, \lambda_p$ control the relative weight of each loss term.

\section{Implementation}

\projname uses a ViT-B/16 backbone and our graph-based Interaction Blocks ($d=512$, $4$ heads).
Our Iterative Refinement Module performs object pose refinement over three iterations.
During training, the model uses ground-truth hand poses as input; at inference time, these are replaced with hand poses estimated by HAMER~\cite{pavlakos2023reconstructinghands3dtransformers}.
We train using Adam with a learning rate of $1\times10^{-4}$ and a batch size of 8.
The appendix provides further training and ablation details.

\section{Experiments}
\label{sec:experiments}

\paragraph{Baselines}
We evaluate \projname against existing methods, noting that prior work focuses exclusively on the hand; to our knowledge, no existing model estimates contact or force on the manipulated object.
While EgoPressure~\cite{zhao2024egopressuredatasethandpressure} addresses 3D hand pressure, its code is not publicly available.
Therefore, we establish two baseline strategies.
\textbf{3D Mesh-Based Comparison:} We utilize the state-of-the-art contact estimation model HACO~\cite{jung2025learningdensehandcontact}.
For contact, we use the publicly available version trained on 14 datasets, following their evaluation protocol on ARCTIC subject \texttt{s05} and H2O \texttt{subject4\_ego}.
To compare force estimation in 3D, we fine-tune the HACO architecture on our augmented ARCTIC training set for 10 epochs, maintaining the original training hyperparameters.
\textbf{2D Image-Based Comparison:} We benchmark against PressureVision~\cite{grady2022pressurevisionestimatinghandpressure}, a model specialized for force that we fine-tune on our training data.
Since it estimates 2D hand pressure maps in image space, we project our 3D per-vertex forces onto the 2D image plane and discretize them into eight pressure bins.
This allows for a fair comparison of hand force localization accuracy within the image domain.

\begin{table*}[b]
\captionsetup{font=footnotesize}
\centering
\scriptsize
\caption{Comparison to PressureVision~\cite{grady2022pressurevisionestimatinghandpressure}, a specialized 2D hand force estimation model, using 2D metrics computed on our projected 3D hand predictions.}
\vspace{-3mm}
\resizebox{\textwidth}{!}{%
\begin{tabular}{@{}l  c@{\hspace{15pt}}
c@{\hspace{5pt}}c@{\hspace{5pt}}c@{\hspace{5pt}}c 
@{\hspace{10pt}}
c@{\hspace{5pt}}c@{\hspace{5pt}}c@{\hspace{5pt}}c@{\hspace{5pt}}c@{}}
& \multicolumn{5}{c}{\textbf{ARCTIC}}
& \multicolumn{5}{c}{\textbf{H2O}} \\
\cmidrule(r{8pt}){2-6} \cmidrule(){7-11}
\textbf{Model} &
\textbf{Prec.} & \textbf{Rec.} & \textbf{F1} & \textbf{IoU} & \textbf{Vol.\ IoU} &
\textbf{Prec.} & \textbf{Rec.} & \textbf{F1} & \textbf{IoU} & \textbf{Vol.\ IoU} \\
\midrule
~~PressureVision        & .101 & .175 & .128 & .068 & ~~6.6 & .004 & .008 & .005 & .003 & 0.2 \\
~~\projname\ w/o IRM   & .186 & .023 & .041 & .021 & ~~1.6 & \textbf{.041} & .013 & .020 & .010 & 0.8 \\
~~\projname             & \textbf{.245} & \textbf{.304} & \textbf{.271} & \textbf{.157} & \textbf{15.4} & .029 & \textbf{.331} & \textbf{.053} & \textbf{.027} & \textbf{2.7} \\
\bottomrule
\end{tabular}
}
\label{tab:pressurevision_comparison}
\end{table*}

\begin{table*}[!h]
\scriptsize
\centering
\caption{Contact and force estimation for HACO\textsuperscript{1} and our models\textsuperscript{2} (Full vs. without Iterative Refinement Module (IRM)). 
\textsuperscript{1}Pretrained on 14 datasets (incl. ARCTIC, H2O). 
\textsuperscript{2}Trained only on ARCTIC train set (excl. s05).}
\vspace{-3mm}
\resizebox{\textwidth}{!}{%
\begin{tabular}{@{}l r
    c@{\hspace{5pt}}c@{\hspace{5pt}}c @{\hspace{10pt}}
    c@{\hspace{5pt}}c@{\hspace{5pt}}c @{\hspace{10pt}}
    c@{\hspace{5pt}}c@{\hspace{5pt}}c@{\hspace{5pt}}c @{\hspace{10pt}}
    c@{\hspace{5pt}}c@{\hspace{5pt}}c@{\hspace{5pt}}c@{}}
& &
\multicolumn{3}{c}{\textbf{\textsc{hand} force}}
& \multicolumn{3}{c}{\textbf{\textsc{object} force}}
& \multicolumn{4}{c}{\textbf{\textsc{hand} contact}}
& \multicolumn{4}{c}{\textbf{\textsc{object} contact}} \\
\cmidrule(r){3-5} \cmidrule(r){6-8}
\cmidrule(r){9-12} \cmidrule(){13-16}
\textbf{Model} &
\textbf{Train/Test} &
\textbf{MAE [N]} & \textbf{RMSE [N]} & \textbf{vIoU} &
\textbf{MAE [N]} & \textbf{RMSE [N]} & \textbf{vIoU} &
\textbf{Prec.} & \textbf{Rec.} & \textbf{F1} & \textbf{IoU} &
\textbf{Prec.} & \textbf{Rec.} & \textbf{F1} & \textbf{IoU} \\
\midrule

\multicolumn{16}{@{}l}{\textbf{evaluation on: ARCTIC s05}} \\[.2em]

~~HACO &
\textbf{in-distribution} &
6.62 & 7.29 & 1.0 &
\multicolumn{3}{c}{\textit{not supported}} &
.120 & \textbf{.886} & \textbf{.196} & \textbf{.113} &
\multicolumn{4}{c}{\textit{not supported}} \\

~~\projname w/o IRM &
\textbf{in-distribution} &
4.80 & 5.94 & 0.7 &
5.01 & 6.03 & 0.7 &
.061 & .092 & .057 & .033 &
.024 & .229 & .038 & .020 \\

~~\projname &
\textbf{in-distribution} &
\textbf{4.03} & \textbf{5.06} & \textbf{2.2} &
\textbf{4.42} & \textbf{5.28} & \textbf{2.0} &
\textbf{.136} & .500 & .190 & .112 &
\textbf{.033} & \textbf{.517} & \textbf{.060} & \textbf{.032} \\

\midrule

\multicolumn{16}{@{}l}{\textbf{evaluation on: H2O s4\_ego}} \\[.2em]

~~HACO &
\textbf{in-distribution} &
6.37 & 7.13 & 0.7 &
\multicolumn{3}{c}{\textit{not supported}} &
\textbf{.127} & \textbf{.831} & \textbf{.198} & \textbf{.117} &
\multicolumn{4}{c}{\textit{not supported}} \\

~~\projname w/o IRM &
\textbf{out-of-distribution} &
\textbf{4.65} & \textbf{5.87} & \textbf{0.8} &
4.70 & 5.35 & 0.4 &
.038 & .040 & .026 & .016 &
.020 & .153 & .030 & .016 \\

~~\projname &
\textbf{out-of-distribution} &
5.16 & 6.62 & 0.6 &
\textbf{3.88} & \textbf{4.18} & \textbf{0.7} &
.064 & .350 & .097 & .055 &
\textbf{.020} & \textbf{.476} & \textbf{.037} & \textbf{.019} \\

\bottomrule
\end{tabular}
}
\label{tab:contact_comparison}
\end{table*}

\paragraph{Evaluation Protocol}
We train exclusively on the ARCTIC dataset, whose objects are strictly articulated and rigid.  
We evaluate on both the ARCTIC test set and H2O.
H2O is excluded from training because its hand and object meshes are less accurate.
Also, many sequences involve interactions with other hands or objects rather than the main object, which could confuse the model.
To evaluate sim-to-real performance, we additionally collected a real-world dataset that captures forces exerted by hands on objects, because such data is highly challenging to obtain and rarely available in existing benchmarks.

\paragraph{Evaluation metrics}
We report precision, recall, and F1 score for contact prediction~\cite{jung2025learningdensehandcontact} along with contact intersection over union (IoU)~\cite{zhao2024egopressuredatasethandpressure, grady2022pressurevisionestimatinghandpressure}.
For force prediction, we report Masked Mean Absolute Error (Masked MAE), Masked Root Mean Squared Error (Masked RMSE), computed within ground-truth contact regions, and volumetric Intersection over Union (vIoU)~\cite{zhao2024egopressuredatasethandpressure, grady2022pressurevisionestimatinghandpressure}.

\paragraph{Ablation Study}
To isolate the contribution of the Iterative Refinement Module, we evaluate a baseline without the recursive object pose update.
Contact and force distributions are predicted directly from the initial template mesh features without further geometric alignment.

\subsection{Results on ARCTIC and H2O}
\label{sec:results_arctic_h2o}

\begin{figure*}[t]
    \centering
    \includegraphics[width=\textwidth]{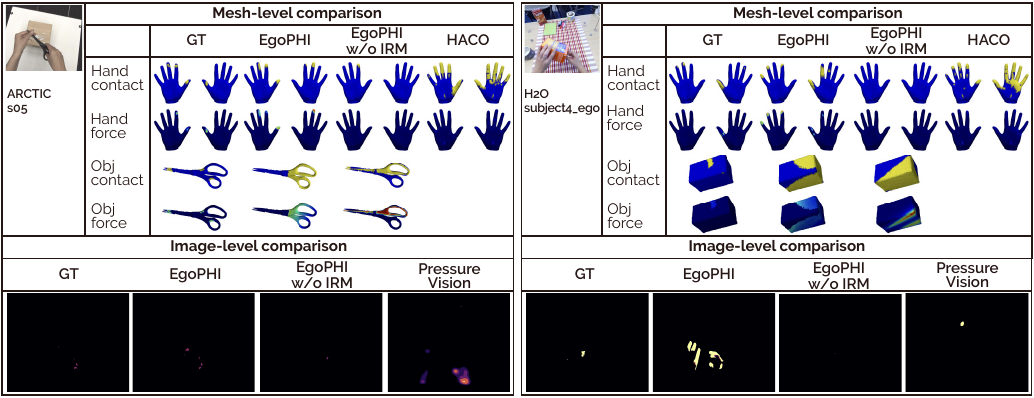}%
    \vspace{-.5ex}%
    \caption{\projname estimates dense per-vertex force and contact for both hands and objects and outperforms the hand-only HACO baseline.
    \projname's projected 2D force distributions outperform PressureVision and align better with the egocentric image.}
    \label{fig:arctic_vis}
\end{figure*}

Tables~\ref{tab:pressurevision_comparison}~and~\ref{tab:contact_comparison} compare \projname's quantitative results with HACO and PressureVision.
We also include an ablated version of \projname without the Iterative Refinement Module.
Evaluation is performed on the ARCTIC and H2O test splits used by the pretrained HACO model.
Section C.1 in the Appendix reports results on the full H2O dataset.
Figure~\ref{fig:arctic_vis} shows qualitative comparisons between \projname, its ablation, HACO, and PressureVision.

\paragraph{Comparison against specialized force baseline}
\projname consistently outperforms PressureVision across both contact and force evaluation metrics.  
While 2D models rely on local appearance cues such as skin blanching, they struggle under the severe occlusions typical of egocentric manipulation. 
\projname leverages 3D geometric reasoning to anchor force predictions to the mesh topology and maintains high localization accuracy even when visual cues are obscured.

\paragraph{Testing on ARCTIC (in-distribution for both)}
\projname surpasses the HACO baseline on force metrics with lower MAE and RMSE and higher Volumetric IoU.
We note that absolute contact scores (F1 < 0.2, IoU < 0.15) reflect the inherent difficulty of the task where ground-truth contact regions are sparse.
For contact regions, \projname achieves higher precision with comparable F1 score and IoU, while HACO attains higher hand-contact recall, likely due to its training on a large, diverse set of 14~hand contact datasets.
For object contact and force estimation, \projname provides the first results of this kind; performance reflects the increased difficulty of localizing interaction on small, occluded object regions, especially when both hand and object poses are estimated.
Overall, \projname produces more physically meaningful force estimates and extends reasoning to object surfaces, a capability not supported by prior methods.

\paragraph{Testing on H2O (out-of-distribution for \projname)}
H2O poses a challenging out-of-distribution setting (unseen objects, scenes, and capture conditions).
Here, \projname remains the only method that produces object contact and force estimates at all, which we consider the primary result.
Our model produces hand forces with lower MAE and RMSE and comparable Volumetric IoU, even though we train it only on ARCTIC.
HACO achieves higher hand contact recall, as expected from its large multi-dataset training.

\paragraph{Ablation of the Iterative Refinement Module (IRM)}  
As shown in Tables~\ref{tab:pressurevision_comparison} and~\ref{tab:contact_comparison}, IRM consistently improves both contact and force estimation across in-distri\-bution (ARCTIC) and out-of-distribution (H2O) datasets.  
In 2D image-level force comparisons, IRM substantially enhances contact localization.
On ARCTIC, IoU increases from 0.021 to 0.157 and volumetric IoU from 1.6 to 15.4 compared to \projname without IRM.
On H2O, IoU rises from 0.010 to 0.027 and volumetric IoU from 0.8 to 2.7.
At the mesh level, the full model achieves consistently better hand and object contact and more accurate object force estimation (IoU and vIoU improve up to 3×).
The only minor exception is H2O hand forces, where the ablated model performs slightly better ($\sim 10\%$).  
These results demonstrate that IRM enforces tighter geometric alignment between hand and object meshes and produces more physically meaningful force predictions.

\begin{figure*}[t]
    \centering
    \includegraphics[width=\textwidth]{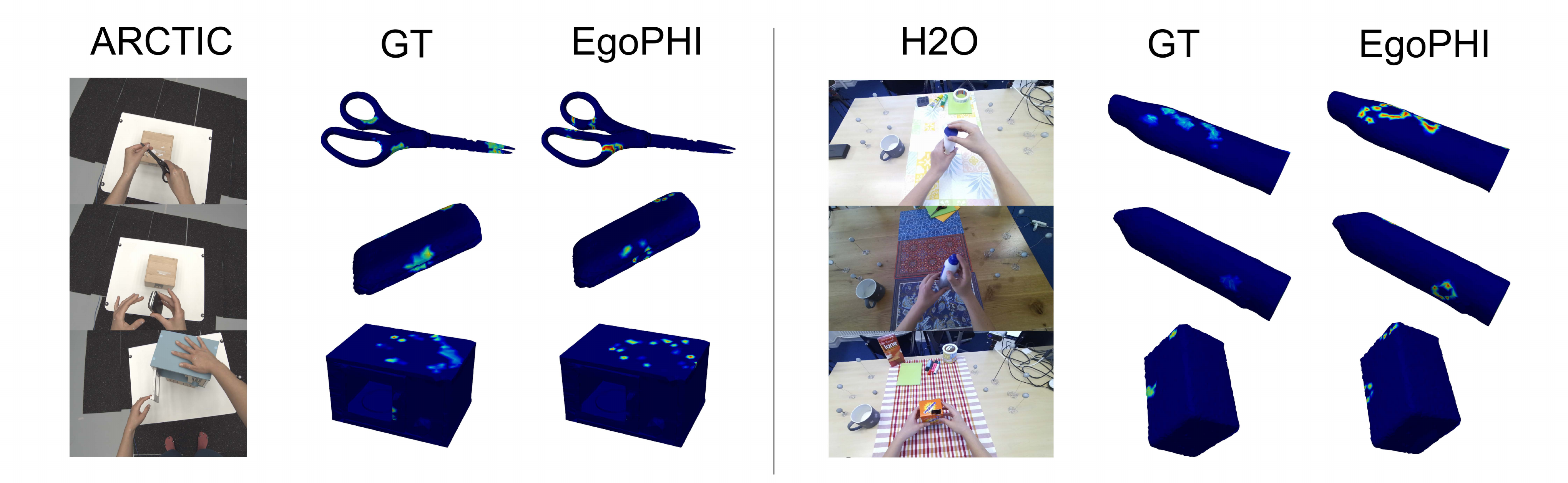}%
    \vspace{-3mm}%
    \caption{Error-isolation analysis with ground-truth poses.}
    \label{fig:GTposes}
\end{figure*}

\paragraph{Pose Refinement Analysis}
Beyond contact metrics, we observe that our iterative refinement significantly improves 3D geometric understanding.
Vertex-level alignment error consistently decreases across iterations, with Mean Per-Vertex Error (MPV) dropping by over $70\%$ (ARCTIC: $64 \rightarrow 19$\,cm, H2O: $48 \rightarrow 10$\,cm). 
Pose correction allows the interaction module to operate more effectively and yields more accurate force and contact predictions even on imperfect initial poses.

\paragraph{Error-Isolation Analysis with Ground-Truth Poses}
To isolate the source of misalignment observed in Figure~\ref{fig:arctic_vis}, we repeat evaluation with ground-truth hand and object poses and bypass pose refinement entirely.
This removes pose-related errors and yields accurate mesh alignment (Figure~\ref{fig:GTposes}).
Results confirm that residual misalignment stems from the difficulty of egocentric object pose estimation under occlusion, rather than from our simulated force labels.

\paragraph{Hand pose domain gap}
\projname is trained with ground-truth hand poses.
At inference time, it relies on HAMER-estimated hand poses~\cite{pavlakos2023reconstructinghands3dtransformers}.
To quantify this train-test domain gap, we fine-tuned \projname using HAMER-estimated hand poses and observed a tradeoff:
hand force prediction improves (MAE: 4.03 $\to$ 3.32~N, RMSE: 5.06 $\to$ 4.08~N, vIoU: 2.2 $\to$ 2.9), since the model adapts to the noisier input distribution at test time.
However, hand contact prediction degrades (F1: 0.190 $\to$ 0.128, Recall: 0.500 $\to$ 0.212), and object force estimation worsens slightly (MAE: 4.42 $\to$ 4.73~N, vIoU: 2.0 $\to$ 1.8), while object contact remains comparable (F1: 0.060 $\to$ 0.061).
We attribute this to a mismatch in our supervision: SOFA supervision is anchored to ground-truth mesh geometry, which becomes inconsistent with HAMER-predicted vertex positions.
Contact labels computed from GT geometry no longer align precisely with the HAMER mesh used at inference, even though the overall force distribution remains learnable.
This highlights an interplay between distribution matching for force regression and geometry-dependent contact supervision.

\begin{figure}[t]
    \centering
    \includegraphics[width=\columnwidth]{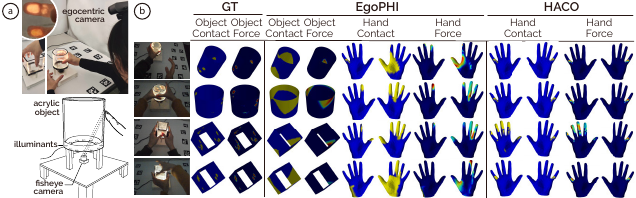}
    \caption{(a)~Apparatus of our real-world contact and force evaluation.
    Participants touched and grasped these two objects, which we constructed such that touch force corresponds to light intensity on the fingers and hands.
    (b)~Qualitative results on real-world force recordings.
    While our experimental apparatus recorded ground-truth contact and force on the object only, we additionally show the estimated contact and force maps on the hands for visual comparison.}%
    \label{fig:setup}
\end{figure}

\subsection{Evaluation on real object contact and force}
\label{sec:eval_real_object}

To evaluate the sim-to-real performance of \projname and validate our force simulation for objects, we designed two touch- and force-sensitive objects: a cube and a cylinder, made from 8\,mm acrylic with polished edges (Figure~\ref{fig:setup}).
Along the bottom of each object we mounted an LED strip (TOPAI 12V COB) and covered it with opaque black tape.
Light is reflected within the acrylic until a finger touches the surface, causing light to escape and illuminate the skin~\cite{han2005low}.
The refractive-index similarity between skin and acrylic causes light to escape into the finger, where it diffuses and reflects.
Because surface residues and fingerprint ridges naturally perturb the optical path, stronger pressure produces brighter reflections and provides an established proxy for force magnitude.

To capture contact areas, we mounted a fisheye camera (8\,MP Sony IMX179 with a near-180° lens) below the object and calibrated camera intrinsics with a checkerboard.
As acrylic has insufficient visual texture for reliable pose detection, we placed ArUco markers on the surface to track the poses of our objects.

We recruited eight participants for the data-capture study.
Each wore a Logitech C920 webcam (78° diagonal FOV) to record egocentric views and interacted with each instrumented object using a set of predefined poses.
These included two-finger grips at multiple locations and force levels; three-, four-, and five-finger configurations; full-hand grasps; and individual touches and presses, with multiple captures per interaction.
Participants performed about ten repetitions per gesture for the left hand, right hand, and both hands.
A typical session lasted under 20~minutes.
Both cameras operated at fixed exposure and 30\,fps.
In total, we collected $\sim$2000~synchronized pairs of egocentric and contact+force frames.

\paragraph{Calibration of light intensity to force}
Before the experiment, we captured calibration data to map light intensity to physical force.
We instrumented a small 8\,mm acrylic panel with LEDs and tape, placed it on the non-illuminated cube above the fisheye camera, and mounted the assembly on a Beurer scale with 1\,g resolution.
An experimenter applied pressure to reach a target weight of 1000\,g and adjusted the fisheye exposure so that the sensor did not saturate.

We then recorded light intensities for a series of controlled force levels.
The experimenter pressed on the panel in 50\,g increments up to 1000\,g (i.e., 9.81\,N) while the fisheye camera captured the reflected light pattern.
For each step, we recorded ten frames, averaged the light intensity over the contact region (excluding a small margin), and converted the measured weights to Newtons.
This lookup table maps optical intensity to force, through which we converted measured optical responses to per-vertex force measurements.

\begin{table}[t]
\scriptsize
\centering
\caption{Real-world evaluation for object contact and force.}%
\vspace{-3mm}%
\begin{tabular}{@{}l @{\hspace{10pt}} l @{\hspace{20pt}} cccc @{\hspace{20pt}} cc @{}}
& &
\multicolumn{4}{c}{\textbf{Object Contact}} &
\multicolumn{2}{c}{\textbf{Object Force [N]}} \\

\cmidrule(r{18pt}){3-6}
\cmidrule(){7-8}

\textbf{Model} & \textbf{Object} &
\textbf{Precision} & \textbf{Recall} & \textbf{F1} &
\textbf{IoU} & \textbf{MAE} & \textbf{RMSE} \\
\midrule

\projname & cylinder &
.114 & .305 & .121 & .067 & 1.35 & 3.24 \\

\projname & cube &
.115 & .316 & .151 & .085 & 0.48 & 1.54 \\

\end{tabular}
\label{tab:realworld_results}
\end{table}

\paragraph{Results}
Table~\ref{tab:realworld_results} shows that \projname can recover real-world object contact and force even though we trained it only on simulated forces (Figure~\ref{fig:setup}).
Force errors are lower than on ARCTIC, which indicates that the model captures object force magnitudes accurately.
This may partly reflect the limited object variety.
Contact prediction is challenging due to small, sparse real contact regions, yet \projname successfully recovers the main contact areas.
These results demonstrate that egocentric vision can infer physically meaningful 3D contact and force distributions on real, non-planar objects.


\begin{table}[b]
\centering
\scriptsize
\setlength{\tabcolsep}{4pt}
\caption{Validation of SOFA simulation against real-world FTIR-based measurements.}
\vspace{-3mm}%
\label{tab:sofa_ftir}
\begin{tabular}{@{}l c c c c@{}}
\textbf{Object} & \textbf{SOFA Force (N)} & \textbf{FTIR Force (N)} & \textbf{Per-frame MAE (N)} & \textbf{Spearman $\rho$} \\
\midrule
Cylinder & $\mu_F = 1.02 \pm 0.59$ & $\mu_F = 1.21 \pm 0.50$ & $0.36 \pm 0.18$ & .760 \\
Cube     & $\mu_F = 1.61 \pm 0.39$ & $\mu_F = 1.67 \pm 0.39$ & $0.25 \pm 0.16$ & .604 \\
\bottomrule
\end{tabular}
\end{table}

\subsection{Validation of Simulated Force Supervision}
\label{sec:simulation_validation}

To verify the fidelity of our SOFA-generated supervision, we compare simulated outputs against our real-world FTIR-instrumented object measurements (Table~\ref{tab:sofa_ftir}). 
SOFA simulation produces mean forces close to real-world FTIR measurements, with low per-frame MAE and strong Spearman correlations.   
These results demonstrate that the simulations provide a reliable and consistent source of force supervision for the rigid objects used in our training and evaluation.

\section{Discussion and Limitations}

Our evaluation shows that egocentric vision, when combined with known object geometry, can effectively infer dense 3D force fields.  
\projname generalizes from simulated supervision to unseen interactions and, to some extent, real-world objects, showing that visual cues support physically meaningful force estimation.

Several limitations remain.  
First, \projname processes frames independently and discards temporal structure. 
Since forces are inherently dynamic, a sequence-aware model could enforce temporal consistency and recover force trajectories that a per-frame approach cannot. 
Second, the model requires the object mesh template as input.
Integrating feed-forward mesh reconstruction models such as SAM3D~\cite{sam3dteam2025sam3d3dfyimages} could make \projname applicable to fully unconstrained egocentric video. 
Third, our force simulation applies uniform stiffness across all objects. 
In our setting, force ratios between vertices are k-independent under a linear penalty, so the spatial distribution of forces is preserved regardless of the chosen constant.
However, this limits accuracy for objects with heterogeneous or compliant material properties. 
Fourth, EgoPHI predicts only normal contact forces, consistent with our simulation and FTIR sensing setup.
Extending to tangential forces would require additional sensing, friction-aware simulation, and learning from weaker visual cues.
A promising direction is post-hoc inference that combines predicted normal forces with object motion via Newton's second law, without extra hardware.
Furthermore, because our model takes only geometry and a single RGB image as input, it lacks object properties (e.g., mass) needed to enforce global force equilibrium.
Finally, the two real-world objects that resolved contact and force in our evaluation were a cube, which is piecewise planar, and a cylinder, which has a developable lateral surface.
Future work should evaluate vision-based contact and force estimation on truly curved 3D objects.
To obtain ground truth on such real-world geometries, our recent fabrication-based approach for retrofitting existing 3D objects with surface-conforming capacitive sensing could be a first step~\cite{ilic2026capsurface}.

\section{Conclusion}
\label{sec:conclusion}

We have presented \projname, a method for estimating physically grounded hand-object interaction from egocentric vision.
\projname estimates dense contact maps and 3D force distributions on both hands and manipulated objects by combining image evidence, object geometry, graph-based interaction reasoning, and iterative pose refinement.
Since dense force annotations are difficult to obtain in natural interaction, we introduced a physics-based simulation pipeline that turns existing hand-object datasets into per-vertex contact and force supervision.

Across simulated, out-of-distribution, and real-world evaluations, \projname shows that egocentric RGB can support meaningful 3D force estimation without tactile sensing at test time.
\projname improves force prediction over prior hand-only and image-space baselines, generalizes to unseen interaction data, and remains the only evaluated approach that estimates force and contact on object surfaces.
Our instrumented-object benchmark further provides dense physical measurements on real 3D objects, offering a step toward validating sim-to-real transfer for force-aware egocentric perception.
At the same time, the remaining errors highlight open challenges in egocentric physical perception, including robust object pose recovery under occlusion, temporal force consistency, and modeling material-dependent interaction.

Our results contribute to a broader shift in egocentric perception from recognizing contact toward estimating the physical interaction it represents.
By predicting forces jointly on hands and objects, \projname adds a force-aware layer to visual hand-object understanding, complementing existing work on contact, pose, and manipulation.
This capability is important because force distributions provide cues about how an object is being held, pressed, stabilized, or controlled, which can support downstream applications such as robot learning from human demonstration, augmented assistance, and embodied scene understanding.

{
\makeatletter
\let\oldurl\url
\renewcommand{\url}[1]{}
\let\olddoi\doi
\renewcommand{\doi}[1]{}
\bibliographystyle{latex/splncs04}
\bibliography{main}

@String(CVPR  = {IEEE Conf. Comput. Vis. Pattern Recog.})

@String(ICCV  = {Int. Conf. Comput. Vis.})

@String(ECCV  = {Eur. Conf. Comput. Vis.})

@String(CVPR  = {CVPR})

@String(ICCV  = {ICCV})

@String(ECCV  = {ECCV})

@misc{huang2022capturinginferringdensefullbody,
      title={Capturing and Inferring Dense Full-Body Human-Scene Contact}, 
      author={Chun-Hao P. Huang and Hongwei Yi and Markus Höschle and Matvey Safroshkin and Tsvetelina Alexiadis and Senya Polikovsky and Daniel Scharstein and Michael J. Black},
      year={2022},
      eprint={2206.09553},
      archivePrefix={arXiv},
      primaryClass={cs.CV},
      url={https://arxiv.org/abs/2206.09553}, 
}

@misc{tripathi2023decodenseestimation3d,
      title={DECO: Dense Estimation of 3D Human-Scene Contact In The Wild}, 
      author={Shashank Tripathi and Agniv Chatterjee and Jean-Claude Passy and Hongwei Yi and Dimitrios Tzionas and Michael J. Black},
      year={2023},
      eprint={2309.15273},
      archivePrefix={arXiv},
      primaryClass={cs.CV},
      url={https://arxiv.org/abs/2309.15273}, 
}

@misc{kim2024contactdiscoveringcomprehensiveaffordance,
      title={Beyond the Contact: Discovering Comprehensive Affordance for 3D Objects from Pre-trained 2D Diffusion Models}, 
      author={Hyeonwoo Kim and Sookwan Han and Patrick Kwon and Hanbyul Joo},
      year={2024},
      eprint={2401.12978},
      archivePrefix={arXiv},
      primaryClass={cs.CV},
      url={https://arxiv.org/abs/2401.12978}, 
}

@misc{wang2025graceestimatinggeometrylevel3d,
      title={GRACE: Estimating Geometry-level 3D Human-Scene Contact from 2D Images}, 
      author={Chengfeng Wang and Wei Zhai and Yuhang Yang and Yang Cao and Zhengjun Zha},
      year={2025},
      eprint={2505.06575},
      archivePrefix={arXiv},
      primaryClass={cs.CV},
      url={https://arxiv.org/abs/2505.06575}, 
}

@inproceedings{han2005low,
  title={Low-cost multi-touch sensing through frustrated total internal reflection},
  author={Han, Jefferson Y},
  booktitle={Proceedings of the 18th annual ACM symposium on User interface software and technology},
  pages={115--118},
  year={2005}
}

@misc{chen2023detectinghumanobjectcontactimages,
      title={Detecting Human-Object Contact in Images}, 
      author={Yixin Chen and Sai Kumar Dwivedi and Michael J. Black and Dimitrios Tzionas},
      year={2023},
      eprint={2303.03373},
      archivePrefix={arXiv},
      primaryClass={cs.CV},
      url={https://arxiv.org/abs/2303.03373}, 
}

@misc{brahmbhatt2020contactposedatasetgraspsobject,
      title={ContactPose: A Dataset of Grasps with Object Contact and Hand Pose}, 
      author={Samarth Brahmbhatt and Chengcheng Tang and Christopher D. Twigg and Charles C. Kemp and James Hays},
      year={2020},
      eprint={2007.09545},
      archivePrefix={arXiv},
      primaryClass={cs.CV},
      url={https://arxiv.org/abs/2007.09545}, 
}

@misc{tse2023s2contactgraphbasednetwork3d,
      title={S$^2$Contact: Graph-based Network for 3D Hand-Object Contact Estimation with Semi-Supervised Learning}, 
      author={Tze Ho Elden Tse and Zhongqun Zhang and Kwang In Kim and Ales Leonardis and Feng Zheng and Hyung Jin Chang},
      year={2023},
      eprint={2208.00874},
      archivePrefix={arXiv},
      primaryClass={cs.CV},
      url={https://arxiv.org/abs/2208.00874}, 
}

@misc{hu2024learningexplicitcontactimplicit,
      title={Learning Explicit Contact for Implicit Reconstruction of Hand-held Objects from Monocular Images}, 
      author={Junxing Hu and Hongwen Zhang and Zerui Chen and Mengcheng Li and Yunlong Wang and Yebin Liu and Zhenan Sun},
      year={2024},
      eprint={2305.20089},
      archivePrefix={arXiv},
      primaryClass={cs.CV},
      url={https://arxiv.org/abs/2305.20089}, 
}

@misc{jung2025learningdensehandcontact,
      title={Learning Dense Hand Contact Estimation from Imbalanced Data}, 
      author={Daniel Sungho Jung and Kyoung Mu Lee},
      year={2025},
      eprint={2505.11152},
      archivePrefix={arXiv},
      primaryClass={cs.CV},
      url={https://arxiv.org/abs/2505.11152}, 
}

@misc{cseke2025picoreconstructing3dpeople,
      title={PICO: Reconstructing 3D People In Contact with Objects}, 
      author={Alpár Cseke and Shashank Tripathi and Sai Kumar Dwivedi and Arjun Lakshmipathy and Agniv Chatterjee and Michael J. Black and Dimitrios Tzionas},
      year={2025},
      eprint={2504.17695},
      archivePrefix={arXiv},
      primaryClass={cs.CV},
      url={https://arxiv.org/abs/2504.17695}, 
}

@misc{cong2025dytactcapturingdynamiccontacts,
      title={DyTact: Capturing Dynamic Contacts in Hand-Object Manipulation}, 
      author={Xiaoyan Cong and Angela Xing and Chandradeep Pokhariya and Rao Fu and Srinath Sridhar},
      year={2025},
      eprint={2506.03103},
      archivePrefix={arXiv},
      primaryClass={cs.CV},
      url={https://arxiv.org/abs/2506.03103}, 
}

@data{ti3dcontact2025,
doi = {10.21227/2smv-hy76},
url = {https://dx.doi.org/10.21227/2smv-hy76},
author = {Zelin Chen and Chenhao Cao and Yiming Ouyang and Hanlu Chen and Hu Jin and Shiwu Zhang},
publisher = {IEEE Dataport},
title = {Ti3D-contact, a high-resolution and whole-body dataset of hand-object contact area based on 3D scanning method},
year = {2024} }

@misc{xie2023visibilityawarehumanobjectinteraction,
      title={Visibility Aware Human-Object Interaction Tracking from Single RGB Camera}, 
      author={Xianghui Xie and Bharat Lal Bhatnagar and Gerard Pons-Moll},
      year={2023},
      eprint={2303.16479},
      archivePrefix={arXiv},
      primaryClass={cs.CV},
      url={https://arxiv.org/abs/2303.16479}, 
}

@misc{wang2024precisionenhancedhumanobjectcontactdetection,
      title={Precision-Enhanced Human-Object Contact Detection via Depth-Aware Perspective Interaction and Object Texture Restoration}, 
      author={Yuxiao Wang and Wenpeng Neng and Zhenao Wei and Yu Lei and Weiying Xue and Nan Zhuang and Yanwu Xu and Xinyu Jiang and Qi Liu},
      year={2024},
      eprint={2412.09920},
      archivePrefix={arXiv},
      primaryClass={cs.CV},
      url={https://arxiv.org/abs/2412.09920}, 
}

@misc{xie2023hmdomarkerlessmultiviewhand,
      title={HMDO: Markerless Multi-view Hand Manipulation Capture with Deformable Objects}, 
      author={Wei Xie and Zhipeng Yu and Zimeng Zhao and Binghui Zuo and Yangang Wang},
      year={2023},
      eprint={2301.07652},
      archivePrefix={arXiv},
      primaryClass={cs.CV},
      url={https://arxiv.org/abs/2301.07652}, 
}

@misc{ehsani2020useforcelukelearning,
      title={Use the Force, Luke! Learning to Predict Physical Forces by Simulating Effects}, 
      author={Kiana Ehsani and Shubham Tulsiani and Saurabh Gupta and Ali Farhadi and Abhinav Gupta},
      year={2020},
      eprint={2003.12045},
      archivePrefix={arXiv},
      primaryClass={cs.CV},
      url={https://arxiv.org/abs/2003.12045}, 
}

@inproceedings{Li_2019,
   title={Estimating 3D Motion and Forces of Person-Object Interactions From Monocular Video},
   url={http://dx.doi.org/10.1109/CVPR.2019.00884},
   DOI={10.1109/cvpr.2019.00884},
   booktitle={2019 IEEE/CVF Conference on Computer Vision and Pattern Recognition (CVPR)},
   publisher={IEEE},
   author={Li, Zongmian and Sedlar, Jiri and Carpentier, Justin and Laptev, Ivan and Mansard, Nicolas and Sivic, Josef},
   year={2019},
   month=jun, pages={8632–8641} }

@inbook{Taheri_2020,
   title={GRAB: A Dataset of Whole-Body Human Grasping of Objects},
   ISBN={9783030585488},
   ISSN={1611-3349},
   url={http://dx.doi.org/10.1007/978-3-030-58548-8_34},
   DOI={10.1007/978-3-030-58548-8_34},
   booktitle={Computer Vision – ECCV 2020},
   publisher={Springer International Publishing},
   author={Taheri, Omid and Ghorbani, Nima and Black, Michael J. and Tzionas, Dimitrios},
   year={2020},
   pages={581–600} }

@misc{hasson2019learningjointreconstructionhands,
      title={Learning joint reconstruction of hands and manipulated objects}, 
      author={Yana Hasson and Gül Varol and Dimitrios Tzionas and Igor Kalevatykh and Michael J. Black and Ivan Laptev and Cordelia Schmid},
      year={2019},
      eprint={1904.05767},
      archivePrefix={arXiv},
      primaryClass={cs.CV},
      url={https://arxiv.org/abs/1904.05767}, 
}

@misc{hampali2021ho3dv3improvingaccuracyhandobject,
      title={HO-3D\_v3: Improving the Accuracy of Hand-Object Annotations of the HO-3D Dataset}, 
      author={Shreyas Hampali and Sayan Deb Sarkar and Vincent Lepetit},
      year={2021},
      eprint={2107.00887},
      archivePrefix={arXiv},
      primaryClass={cs.CV},
      url={https://arxiv.org/abs/2107.00887}, 
}

@misc{zhao2024egopressuredatasethandpressure,
      title={EgoPressure: A Dataset for Hand Pressure and Pose Estimation in Egocentric Vision}, 
      author={Yiming Zhao and Taein Kwon and Paul Streli and Marc Pollefeys and Christian Holz},
      year={2024},
      eprint={2409.02224},
      archivePrefix={arXiv},
      primaryClass={cs.CV},
      url={https://arxiv.org/abs/2409.02224}, 
}

@misc{zhang2024forcephysicsawarehumanobjectinteraction,
      title={FORCE: Physics-aware Human-object Interaction}, 
      author={Xiaohan Zhang and Bharat Lal Bhatnagar and Sebastian Starke and Ilya Petrov and Vladimir Guzov and Helisa Dhamo and Eduardo Pérez-Pellitero and Gerard Pons-Moll},
      year={2024},
      eprint={2403.11237},
      archivePrefix={arXiv},
      primaryClass={cs.CV},
      url={https://arxiv.org/abs/2403.11237}, 
}

@INPROCEEDINGS{7298898,
  author={Tu-Hoa Pham and Kheddar, Abderrahmane and Qammaz, Ammar and Argyros, Antonis A.},
  booktitle={2015 IEEE Conference on Computer Vision and Pattern Recognition (CVPR)}, 
  title={Towards force sensing from vision: Observing hand-object interactions to infer manipulation forces}, 
  year={2015},
  volume={},
  number={},
  pages={2810-2819},
  doi={10.1109/CVPR.2015.7298898}}

@misc{grady2022pressurevisionestimatinghandpressure,
      title={PressureVision: Estimating Hand Pressure from a Single RGB Image}, 
      author={Patrick Grady and Chengcheng Tang and Samarth Brahmbhatt and Christopher D. Twigg and Chengde Wan and James Hays and Charles C. Kemp},
      year={2022},
      eprint={2203.10385},
      archivePrefix={arXiv},
      primaryClass={cs.CV},
      url={https://arxiv.org/abs/2203.10385}, 
}

@misc{grady2024pressurevisionestimatingfingertippressure,
      title={PressureVision++: Estimating Fingertip Pressure from Diverse RGB Images}, 
      author={Patrick Grady and Jeremy A. Collins and Chengcheng Tang and Christopher D. Twigg and Kunal Aneja and James Hays and Charles C. Kemp},
      year={2024},
      eprint={2301.02310},
      archivePrefix={arXiv},
      primaryClass={cs.CV},
      url={https://arxiv.org/abs/2301.02310}, 
}

@misc{fan2023arcticdatasetdexterousbimanual,
      title={ARCTIC: A Dataset for Dexterous Bimanual Hand-Object Manipulation}, 
      author={Zicong Fan and Omid Taheri and Dimitrios Tzionas and Muhammed Kocabas and Manuel Kaufmann and Michael J. Black and Otmar Hilliges},
      year={2023},
      eprint={2204.13662},
      archivePrefix={arXiv},
      primaryClass={cs.CV},
      url={https://arxiv.org/abs/2204.13662}, 
}

@misc{kwon2021h2ohandsmanipulatingobjects,
      title={H2O: Two Hands Manipulating Objects for First Person Interaction Recognition}, 
      author={Taein Kwon and Bugra Tekin and Jan Stuhmer and Federica Bogo and Marc Pollefeys},
      year={2021},
      eprint={2104.11181},
      archivePrefix={arXiv},
      primaryClass={cs.CV},
      url={https://arxiv.org/abs/2104.11181}, 
}

@misc{liu2024hoi4d4degocentricdataset,
      title={HOI4D: A 4D Egocentric Dataset for Category-Level Human-Object Interaction}, 
      author={Yunze Liu and Yun Liu and Che Jiang and Kangbo Lyu and Weikang Wan and Hao Shen and Boqiang Liang and Zhoujie Fu and He Wang and Li Yi},
      year={2024},
      eprint={2203.01577},
      archivePrefix={arXiv},
      primaryClass={cs.CV},
      url={https://arxiv.org/abs/2203.01577}, 
}

@misc{chao2021dexycbbenchmarkcapturinghand,
      title={DexYCB: A Benchmark for Capturing Hand Grasping of Objects}, 
      author={Yu-Wei Chao and Wei Yang and Yu Xiang and Pavlo Molchanov and Ankur Handa and Jonathan Tremblay and Yashraj S. Narang and Karl Van Wyk and Umar Iqbal and Stan Birchfield and Jan Kautz and Dieter Fox},
      year={2021},
      eprint={2104.04631},
      archivePrefix={arXiv},
      primaryClass={cs.CV},
      url={https://arxiv.org/abs/2104.04631}, 
}

@InProceedings{Cao2021,
  title = {Reconstructing Hand-Object Interactions in the Wild},
  author = {Zhe Cao and Ilija Radosavovic and Angjoo Kanazawa and Jitendra Malik},
  booktitle = {ICCV},
  year = {2021}
}

@INPROCEEDINGS{hampali2020honnotate,
	      title={HOnnotate: A method for 3D Annotation of Hand and Object Poses},
          author={Shreyas Hampali and Mahdi Rad and Markus Oberweger and Vincent Lepetit},
          booktitle = {CVPR},
      year = {2020}
         }

@misc{pavlakos2023reconstructinghands3dtransformers,
      title={Reconstructing Hands in 3D with Transformers}, 
      author={Georgios Pavlakos and Dandan Shan and Ilija Radosavovic and Angjoo Kanazawa and David Fouhey and Jitendra Malik},
      year={2023},
      eprint={2312.05251},
      archivePrefix={arXiv},
      primaryClass={cs.CV},
      url={https://arxiv.org/abs/2312.05251}, 
}

@article{Zhu2025ForcesFF,
  title={Forces for free: Vision-based contact force estimation with a compliant hand},
  author={Yifan Zhu and Mei Hao and Xupeng Zhu and Quentin Bateux and Alex Wong and Aaron M. Dollar},
  journal={Science robotics},
  year={2025},
  volume={10 103},
  pages={
          eadq5046
        },
  url={https://api.semanticscholar.org/CorpusID:279599740}
}

@misc{yang2024visionbasedforceestimationminimally,
      title={Vision-Based Force Estimation for Minimally Invasive Telesurgery Through Contact Detection and Local Stiffness Models}, 
      author={Shuyuan Yang and My H. Le and Kyle R. Golobish and Juan C. Beaver and Zonghe Chua},
      year={2024},
      eprint={2403.18172},
      archivePrefix={arXiv},
      primaryClass={cs.RO},
      url={https://arxiv.org/abs/2403.18172}, 
}

@incollection{faure:hal-00681539,
  TITLE = {{SOFA: A Multi-Model Framework for Interactive Physical Simulation}},
  AUTHOR = {Faure, Fran{\c c}ois and Duriez, Christian and Delingette, Herv{\'e} and Allard, J{\'e}r{\'e}mie and Gilles, Benjamin and Marchesseau, St{\'e}phanie and Talbot, Hugo and Courtecuisse, Hadrien and Bousquet, Guillaume and Peterlik, Igor and Cotin, St{\'e}phane},
  URL = {https://inria.hal.science/hal-00681539},
  BOOKTITLE = {{Soft Tissue Biomechanical Modeling for Computer Assisted Surgery}},
  EDITOR = {Yohan Payan},
  PUBLISHER = {{Springer}},
  SERIES = {Studies in Mechanobiology, Tissue Engineering and Biomaterials},
  VOLUME = {11},
  PAGES = {283-321},
  YEAR = {2012},
  MONTH = Jun,
  DOI = {10.1007/8415\_2012\_125},
  HAL_ID = {hal-00681539},
  HAL_VERSION = {v1},
}

@misc{yu2025dynamicreconstructionhandobjectinteraction,
      title={Dynamic Reconstruction of Hand-Object Interaction with Distributed Force-aware Contact Representation}, 
      author={Zhenjun Yu and Wenqiang Xu and Pengfei Xie and Yutong Li and Brian W. Anthony and Zhuorui Zhang and Cewu Lu},
      year={2025},
      eprint={2411.09572},
      archivePrefix={arXiv},
      primaryClass={cs.CV},
      url={https://arxiv.org/abs/2411.09572}, 
}

@ARTICLE{8085141,
  author={Pham, Tu-Hoa and Kyriazis, Nikolaos and Argyros, Antonis A. and Kheddar, Abderrahmane},
  journal={IEEE Transactions on Pattern Analysis and Machine Intelligence}, 
  title={Hand-Object Contact Force Estimation from Markerless Visual Tracking}, 
  year={2018},
  volume={40},
  number={12},
  pages={2883-2896},
  doi={10.1109/TPAMI.2017.2759736}}

@misc{jiang2024contactimplicitmodelpredictivecontrol,
      title={Contact-Implicit Model Predictive Control for Dexterous In-hand Manipulation: A Long-Horizon and Robust Approach}, 
      author={Yongpeng Jiang and Mingrui Yu and Xinghao Zhu and Masayoshi Tomizuka and Xiang Li},
      year={2024},
      eprint={2402.18897},
      archivePrefix={arXiv},
      primaryClass={cs.RO},
      url={https://arxiv.org/abs/2402.18897}, 
}

@article{SMPL:2015,
      author = {Loper, Matthew and Mahmood, Naureen and Romero, Javier and Pons-Moll, Gerard and Black, Michael J.},
      title = {{SMPL}: A Skinned Multi-Person Linear Model},
      journal = {ACM Trans. Graphics (Proc. SIGGRAPH Asia)},
      month = oct,
      number = {6},
      pages = {248:1--248:16},
      publisher = {ACM},
      volume = {34},
      year = {2015}
    }

@inproceedings{SMPL-X:2019,
  title = {Expressive Body Capture: {3D} Hands, Face, and Body from a Single Image},
  author = {Pavlakos, Georgios and Choutas, Vasileios and Ghorbani, Nima and Bolkart, Timo and Osman, Ahmed A. A. and Tzionas, Dimitrios and Black, Michael J.},
  booktitle = {Proceedings IEEE Conf. on Computer Vision and Pattern Recognition (CVPR)},
  pages     = {10975--10985},
  year = {2019}
}

@article{MANO:SIGGRAPHASIA:2017,
      title = {Embodied Hands: Modeling and Capturing Hands and Bodies Together},
      author = {Romero, Javier and Tzionas, Dimitrios and Black, Michael J.},
      journal = {ACM Transactions on Graphics, (Proc. SIGGRAPH Asia)},
      volume = {36},
      number = {6},
      series = {245:1--245:17},
      month = nov,
      year = {2017},
      month_numeric = {11}
  }

@article{sam3dteam2025sam3d3dfyimages,
      title={SAM 3D: 3Dfy Anything in Images}, 
      author={SAM 3D Team and Xingyu Chen and Fu-Jen Chu and Pierre Gleize and Kevin J Liang and Alexander Sax and Hao Tang and Weiyao Wang and Michelle Guo and Thibaut Hardin and Xiang Li and Aohan Lin and Jiawei Liu and Ziqi Ma and Anushka Sagar and Bowen Song and Xiaodong Wang and Jianing Yang and Bowen Zhang and Piotr Dollár and Georgia Gkioxari and Matt Feiszli and Jitendra Malik},
      year={2025},
      eprint={2511.16624},
      archivePrefix={arXiv},
      primaryClass={cs.CV},
      url={https://arxiv.org/abs/2511.16624}, 
}

@inproceedings{ilic2026capsurface,
  author = {Andjela Ilic and Junpeng Gao and Zhipeng Li and Yijing Jiang and Rachel Schuchert and Manuel Meier and Philipp Herholz and Christian Holz},
  title = {Retrofitting Existing 3D Objects with Surface-Conforming Capacitive Sensing},
  year = {2026},
  publisher = {Association for Computing Machinery},
  address = {New York, NY, USA},
  booktitle = {ACM SIGGRAPH 2026 Conference Papers},
  location = {Los Angeles, CA, USA},
  series = {SIGGRAPH '26}
}

@inproceedings{streli2021capcontactsuperresolution,
  title={{CapContact}: Super-resolution Contact Areas from Capacitive Touchscreens},
  author={Streli, Paul and Holz, Christian},
  booktitle={Proceedings of the ACM CHI Conference on Human Factors in Computing Systems},
  year={2021},
  publisher={Association for Computing Machinery},
  address={New York, NY, USA}
}

@inproceedings{schuchert2026capfm,
  title={{CapFM}: A Foundation Model for Capacitive Touch Sensing},
  author={Schuchert, Rachel and Holz, Christian},
  booktitle={Proceedings of the ACM Symposium on User Interface Software and Technology (UIST)},
  year={2026},
  publisher={Association for Computing Machinery},
  address={New York, NY, USA}
}

@inproceedings{streli2023structuredlightspeckle,
  title={Structured Light Speckle: Joint egocentric depth estimation and low-latency contact detection via remote vibrometry},
  author={Streli, Paul and Jiang, Jiaxi and Rossie, Juliete and Holz, Christian},
  booktitle={Proceedings of the 36th Annual ACM Symposium on User Interface Software and Technology (UIST)},
  year={2023},
  publisher={Association for Computing Machinery},
  address={New York, NY, USA}
}

@inproceedings{streli2024touchinsightuncertaintyaware,
  title={{TouchInsight}: Uncertainty-aware Rapid Touch and Text Input for Mixed Reality from Egocentric Vision},
  author={Streli, Paul and Richardson, Mark and Botros, Fadi and Ma, Shugao and Wang, Robert and Holz, Christian},
  booktitle={Proceedings of the 37th Annual ACM Symposium on User Interface Software and Technology (UIST)},
  year={2024},
  publisher={Association for Computing Machinery},
  address={New York, NY, USA},
  doi={10.1145/3654777.3676330}
}
\let\url\oldurl
\let\doi\olddoi
\makeatother
}

\end{document}